\documentclass{llncs}
\usepackage{graphicx}
\usepackage{multicol}
\usepackage{color}
\usepackage{listings}

\definecolor{gris}{gray}{0.5}

\lstdefinelanguage{atl}{
	morekeywords={rule,from,to},
	sensitive=false,
        linewidth=\linewidth,
	morecomment=[l]{--},
	morestring=[b]',
}
\lstdefinelanguage{scomma}{
	morekeywords={int,in,intersect,card,class,forall,if,then,else,enum,
          constraint,set,main},
	sensitive=false,
	morecomment=[l]{//},
	morecomment=[s]{/*}{*/},
	morestring=[b]',
}
\lstdefinestyle{scma}{
	language=scomma,
	mathescape=true, 
	tabsize=3,
	basicstyle=\scriptsize, 
	keywordstyle=\color{black}\textbf,
        numbers={left},
        breaklines=true,
        xleftmargin=25pt,
	commentstyle=\color{gris}\scriptsize\textrm, 
	stringstyle=\ttfamily, 
	showstringspaces=false
}
\lstdefinelanguage{Ltcs}{
	morekeywords={template,refersTo,separator,context,importContext,lookIn,
        addToContext,main,isDefined},
	sensitive=false,
	morecomment=[l]{--},
	morestring=[b]',
}
\lstdefinestyle{tcs}{
	language=Ltcs,
	mathescape=true, 
	tabsize=3,
	basicstyle=\scriptsize, 
	keywordstyle=\color{black}\textbf,
        numbers={left},
        xleftmargin=25pt,
        linewidth=0.9\linewidth,
        breaklines=true,
	commentstyle=\color{gris}\footnotesize\textrm, 
	stringstyle=\ttfamily, 
	showstringspaces=false
}
\lstdefinelanguage{eclipse}{
	morekeywords={intset,for,dim,param,do},
	sensitive=false,
	morecomment=[l]{//},
	morecomment=[s]{/*}{*/},
	morestring=[b]',
}
\lstdefinestyle{ecl}{
	language=eclipse,
	mathescape=true, 
	tabsize=3,
	basicstyle=\scriptsize, 
	keywordstyle=\color{black}\textbf,
        numbers={left},
        xleftmargin=25pt,
        linewidth=0.9\linewidth,
        breaklines=true,
	commentstyle=\color{gris}\footnotesize\textrm, 
	stringstyle=\ttfamily, 
	showstringspaces=false
}

\newcommand{\sCOMMA} {\textsf{\footnotesize{s-COMMA}}}

\newcommand{\code}[1]{\texttt{\small{#1}}}

\newcommand{\Eclipse} {ECL$^{i}$PS$^{e}$}

\title{Using ATL to define advanced and flexible constraint model transformations}
\author{Rapha\"el Chenouard\inst{1} \and Laurent Granvilliers\inst{1}
  \and Ricardo Soto\inst{1,2} }
\institute{LINA, CNRS, Universit\'e de Nantes, France\\
\and
Escuela de Ingenier{\'\i}a Inform\'atica\\ Pontificia Universidad
Cat\'olica de Valpara{\'\i}so, Chile\\
{\texttt{$\{$raphael.chenouard,laurent.granvilliers,ricardo.soto$\}$@univ-nantes.fr}}}

\begin{document}
\maketitle

\begin{abstract}
Transforming constraint models is an important task in recent
constraint programming systems. User-understandable models are defined
during the modeling phase but rewriting or tuning them is mandatory to
get solving-efficient models. We propose a new architecture allowing
to define bridges between any (modeling or solver) languages and to
implement model optimizations. This architecture follows a
model-driven approach where the constraint modeling process is seen
as a set of model transformations. Among others, an interesting
feature is the definition of transformations as concept-oriented
rules, i.e. based on types of model elements where the types
are organized into a hierarchy called a metamodel.
\end{abstract}

\section{Introduction}\label{sec:intro}
Constraint programming (CP) systems must combine a modeling language
and a solving engine. The modeling language is used to represent
problems with variables, constraints, or statements. The solving
engine computes assignments of variables satisfying the constraints by
exploring and pruning the space of potential solutions. This paper
considers the constraint modeling process as constraint model
transformations between arbitrary modeling or solver languages. It
follows several important consequences on the architecture of systems
and user practices.

Constraint programming languages are rich, combining common constraint
domains, e.g. integer constraints or linear real constraints, with
global constraints like \texttt{alldifferent}, and even statements
like \texttt{if-then-else} or \texttt{forall}. Moreover the spectrum
of syntaxes is large, ranging from computer programming languages like
Java or Prolog to high-level languages intended to be more
human-comprehensible. This may be contrasted with the existence of a
standard language in the field of mathematical programming, which
improves model sharing, writing and understanding. The quest of a
standard CP language is a recent thread, dating back to the talk of
Puget~\cite{PugetCP04}. Another important concern is to employ the
best solving technology for a given model. As a consequence, a new
kind of architecture emerged. The key idea is to map models written
with a high-level CP language to many solvers. For instance within the
G12 project, MiniZinc~\cite{Nethercote2007} is intended to be a
standard modeling language, and Cadmium~\cite{Brand2008} is able to
map MiniZinc models to a set of
solvers. Essence~\cite{FrischIJCAI2007} is another CP platform
offering an high level modeling language refining Essence
specifications to Essence' models using
Conjure~\cite{FrischIJCAI2005}. Then hand-written translators can
generate models for several different solvers. The role of a mapping
tool is to bridge modeling and solver languages and to optimize models
for improving the solving process. Cadmium is based on Constraint
Handling Rules \cite{CHRBook2009} and is the the closest CP platform
from our model-driven approach.

In our approach, we suppose that any CP language can be chosen at the
modeling phase. In fact, finding a standard language is hard and
existing languages have their own features. It then becomes
necessary to define mappings between any (pure modeling or solver)
languages. This is just the first goal of the new architecture for
constraint model transformations defined in the sequel. It follows
many advantages:
\begin{itemize}
\item Any user may choose its favourite modeling language and the known
  best solving technology for a given problem provided that the
  transformation between languages is implemented.

\item It may be easy to create a collection of benchmarks for a given
  language from different source languages. This feature may speed up
  prototyping of one solver, avoiding hand rewriting of problems into
  the solver language.

\item A given problem may be handled using different solving
  technologies. Users may not have to play with solver languages.
\end{itemize}
To this end, we define a generic and flexible pivot model
(i.e. an intermediate model) to which any language is
mapped. Considering a new language in this framework only requires a
parser and a generally simple transformation to the pivot model.

The second goal is to define refactoring operations and optimizations
of constraint models using declarative rules. Implementing
them over pivot models guarantees the independence from external
languages. In other words every operation is implemented once, by
means of a so-called concept-oriented rule. In our model engineering
approach the elements of models are specified within metamodels, which
can be seen as a hierarchy of concepts or types. The rules are able to
filter models according to these types, which may be more powerful than
syntax-oriented rules.

The third goal is to apply the best transformations for given solving
technologies. For instance, a matrix with a few non null elements
could be transformed into a sparse matrix when using a linear algebra
package. The selection of transformation steps is implemented as a
sequential procedure, applying transformations until at least pivot
models fit the structure requirements of the target language.

This architecture has been fully implemented using a model-driven
engineering (MDE) approach~\cite{OMG_MDA}. MDE tools enable us to
separate the grammar concerns from modeling concepts using dedicated
tools and languages like TCS~\cite{Jouault2006TCS} and
ATL~\cite{Jouault2006ATL,JouaultATL2008}. The main advantage is that we can reason
about concepts and their relations through a metamodel.
Transformations are specified by defining matchings between concepts
at the metamodel level of abstraction. Thus, grammar concerns are
relegated into the foreground, while concepts processing becomes the
major task.

With respect to previous works, e.g.~\cite{Chenouard2008}, the new
architecture gives more freedom in constraint modeling. \sCOMMA{} is
not always the source modeling language and refactoring steps can be
chosen. Thus, users can play with any modeling language, until it is
mapped to our platform. Dealing with a
solver does not require to manipulate its language. Moreover, handling
a new language or a new transformation in the system requires a few
work. The main limitation of our approach is that only the modeling
fragments of languages can be processed i.e., the declarative part. It
is not possible to partially execute a computer program that builds the
constraint store.

This paper is organized as follows. Section~\ref{sec:overview}
presents an overview of our general transformation framework. Next section
introduces the metamodels of two CP languages illustrated on a well-known
problem. The pivot metamodel and the transformation rules are introduced
in Section~\ref{sec:intermediate}. Section~\ref{sec:bidirectional}
presents the whole model-driven process including the possibility of
selecting relevant mappings. The related work and a conclusion follow.

\section{The Model-Driven Transformation Framework}\label{sec:overview}

\vspace*{-2mm}
\begin{figure}[tbp]
\begin{center}
\includegraphics[width=0.70\linewidth,bb=136 444 538 655]
                {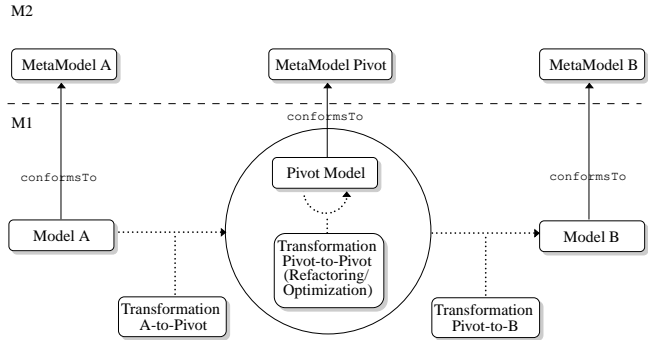}
\caption{General transformation framework.}\label{fig:framework-mda}
\end{center}
\vspace*{-2mm}
\end{figure}

Figure~\ref{fig:framework-mda} depicts the architecture of our
model-driven transformation framework, which is classically divided in
two layers M1 and M2~\cite{OMG_MDA}. M1 holds the models representing
constraint problems and M2 defines the semantic of M1 through
metamodels. Metamodels describe the concepts appearing in models,
e.g. constraint, variable, or domain, and the relations among these
concepts, e.g. inheritance, composition, or association.
In this framework, transformation rules are defined
to perform a complete translation in three main steps: translation
from source model A to the pivot model, refactoring/optimization on
the pivot model, and translation from the pivot model to target model
B.  Models A and B may be defined through any CP languages. The pivot
model may be refined several times in order to adapt it to the desired
target model (see Section~\ref{sec:intermediate}).

A main feature resulting from a model-driven engineering approach is
that transformation rules operate on the metamodel concepts. For
instance, unrolling a \texttt{forall} loop is implemented once over
the \texttt{forall} concept, which is independent from the many
syntaxes of \texttt{forall} in CP languages. In fact, no grammar
specification is required for the pivot model. Syntax specifications
of CP languages must be defined separately using specific tools
achieving text-to-model or model-to-text mappings
like TCS~\cite{Jouault2006TCS}, which implement both tasks.

\section{A Motivating Example}\label{sec:CPMM}
In this section, we consider two CP languages, and we motivate the
needs and the means for implementing transformations between them.

\begin{figure}[tbp]
 \includegraphics [width=.95\linewidth] 
 {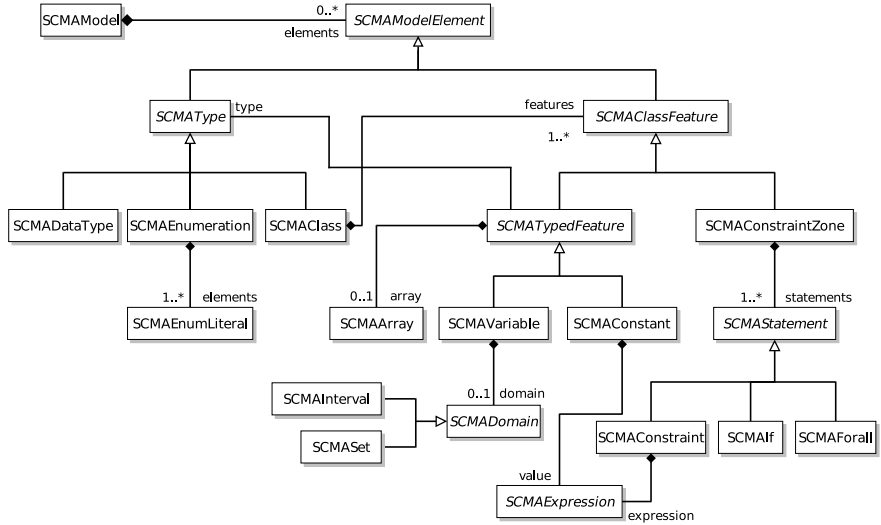} 
\caption{Extract of the \sCOMMA{} metamodel.}\label{fig:scomma-metamodel}
\vspace*{-2mm}
\end{figure}

\begin{figure}[tbp]
 \includegraphics [width=.95\linewidth] 
 {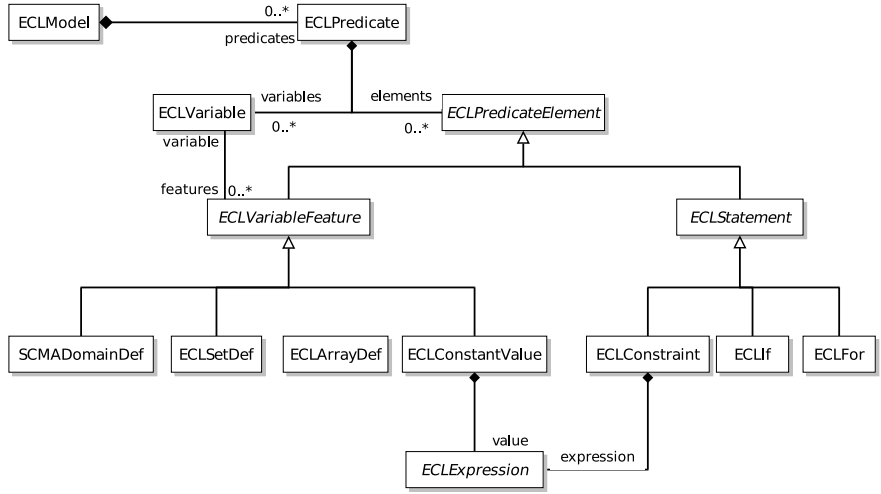} 
\caption{Extract of the \Eclipse{} metamodel.}\label{fig:eclipse-metamodel}
\vspace*{-2mm}
\end{figure}

\Eclipse{}~\cite{Wallace97eclipse} is chosen as a leading constraint
logic programming system. \sCOMMA~\cite{SotoICTAI2007} is an
object-oriented constraint language developed in our team. Their
metamodels are partially depicted in Figure~\ref{fig:scomma-metamodel}
and ~\ref{fig:eclipse-metamodel}
using UML class diagram notation. The roots of these hierarchies are
equivalent, such that the model concept represents the complete
constraint problem to be processed.

In \sCOMMA{}, a model is composed of a collection of model elements. A 
model element is either an enumeration, or a class, or a constant. Each 
class is composed of a set of class features which 
can be specialized in variables, constant or constraint
zones. Variable with a type defined as a class is an
object. Constraint zones are used to group constraints and
other statements such as conditionals and loops. The concepts of
global constraints and optimization objective are not shown here, but
can be also defined. The concept of expressions are not detailed in
this paper since it is based on classical operatored expressions using
boolean, set and arithmetic operators.

In the \Eclipse{} metamodel, we propose to define a model as a
collection of predicates holding predicate elements and
variables. Predicate elements are variable features or
statements. Variables features is either a constant value 
assignment, a domain definition, an array or a set definition related
to a variable. In fact, we consider that variables are implicitly
declared through their features.

Considering the well-known problem of the social golfers,
Figure~\ref{fig:scomma-golfers} and \ref{fig:eclipse-golfers} show two
versions of the same problem using \sCOMMA{} and \Eclipse{}
languages. This problem considers a group of $n=g\times s$ golfers
that wish to play golf each week, arranged into $g$ groups of $s$
golfers, the problem is to find a playing schedule for $w$ weeks such
that no two golfers play together more than once.

\begin{figure}[htpb]
\begin{lstlisting}[style=scma,multicols=2,linewidth=0.9\linewidth]
// Data file
enum Name := {a,b,c,d,e,f,g,h,i};
int s := 3; //size of groups
int w := 4; //number of weeks
int g := 3; //groups per week

// Model file
main class SocialGolfers {
 Week weeks[w];
 constraint differentGroups {
  forall(w1 in 1..w) {
   forall(w2 in w1+1..w) {
    card(weeks[w1].groups[g1].players intersect weeks[w2].groups[g2].players)<= 1;
   }
  }
 }
}
class Week {
 Group groups[g];
 constraint playOncePerWeek { 
  forall(g1 in 1..g) {
   forall(g2 in g1+1..g) {
    card(groups[g1].players intersect groups[g2].players) = 0;
   }
  }
 }
}
class Group {
 Name set players;
 constraint groupSize {
  card(players) = s;
 }
}
\end{lstlisting}
\caption{The social golfers problem expressed in \sCOMMA.}\label{fig:scomma-golfers}
\end{figure}

The \sCOMMA{} model is divided in a data file and a model file. The
data file contains the golfer names encoded as an \code{Enum} concept
at line 1 and the problem dimensions defined by means of 
constants (size of groups, number of weeks, and groups per 
week). The model file represents the generic
social golfers problem using the \code{Model} concept. The problem
structure is captured by the three classes \code{SocialGolfers},
\code{Group}, and \code{Week}, which are conformed to the \code{Class}
concept. The \code{Group} class owns the \code{players} attribute
corresponding to a set of golfers playing together, each golfer being
identified by a name given in the enumeration from the data file. In
this class, the constraint zone \code{groupSize} (lines 30 to 32)
restricts the size of the golfers group. The \code{Week} class has an
array of \code{Group} objects and the constraint zone
\code{playOncePerWeek} ensures that each golfer takes part of a unique
group per week.  Finally, the \code{SocialGolfers} class has an array
of \code{Week} objects and the constraint zone \code{differentGroups}
states that each golfer never plays two times with the same golfer
throughout the considered weeks.

\begin{figure}[htpb]
\begin{lstlisting}[style=ecl,multicols=2,linewidth=0.9\linewidth,escapechar=\%]
socialGolfers(L):-
 S %\$%= 3,
 W %\$%= 4,
 G %\$%= 3,
 intsets(WEEKS_GROUPS_PLAYERS,12,1,9),
 L = WEEKS_GROUPS_PLAYERS,

 (for(W1,1,W), param(L,W,G) do
  (for(W2,W1+1,W), param(L,G,W1) do
   (for(G1,1,G), param(L,G,W1,W2) do
    (for(G2,1,G), param(L,G,W1,W2,G1) do
     V1 is G*(W1-1)+G1,
     nth(V2,V1,L),
     V3 is G*(W2-1)+G2,
     nth(V4,V3,L),
     #(V2 /\ V4, V5),V5 %\$%=< 1
    )
   )
  )
 ),
 (for(WEEKS,1,W),param(L,G) do
  (for(GROUPS,1,G), param(L,S,W,G,WEEKS) do
   V6 is G*(WEEKS-1)+GROUPS,
   nth(V7,V6,L), 
   #(V7, V8), V8 %\$%= S,

   (for(G1,1,G),param(L,G,WEEKS) do
    (for(G2,G1+1,G),param(L,G,WEEKS,G1) do
     V9 is G*(WEEKS-1)+G1,
     nth(V10,V9,L),
     V11 is G*(WEEKS-1)+G2,
     nth(V12,V11,L),
     #(V10 /\ V12, 0)
    )
   )
  )
 ),

 label_sets(L).
\end{lstlisting}
\caption{The social golfers problem expressed in \Eclipse.}\label{fig:eclipse-golfers}
\end{figure}

Figure~\ref{fig:eclipse-golfers} depicts the \Eclipse{} model
resulting from an automatic transformation of the previous \sCOMMA{}
model. The problem is now encoded as a single predicate whose body is
a sequence of atoms. The sequence is made of the problem dimensions,
the list of constrained variables \code{L}, and three statements
resulting from the transformation of the three \sCOMMA{} classes. It
turns out that parts of both models are similar. This is due to the
sharing of concepts in the underlying metamodels, for instance
constants, \texttt{forall} statements, or constraints. However, the
syntaxes are different and specific processing may be required. For
instance, the \texttt{forall} statement of \Eclipse{} needs the
\code{param} keyword to declare parameters defined outside of the
current scope, e.g. the number of groups \texttt{G}.

The treatment of objects is more subtle since they must not
participate to \Eclipse{} models. Many mapping strategies may be
devised, for instance mapping objects to
predicates~\cite{SotoICTAI2007}. Another mapping strategy is used
here, which consists in removing the object-based problem
structure. Flattening the problem requires visiting the many classes
through their inheritance and composition relations.  A few problems
to be handled are described as follows. Important changes on the
attributes may be noticed. For example, the \code{weeks} array of
\code{Week} objects defined at line 9 in
Figure~\ref{fig:scomma-golfers} is refactored and transformed to the
\code{WEEKS\_GROUPS\_PLAYERS} flat list stated at line 5 in
Figure~\ref{fig:eclipse-golfers}. It may be possible to insert new
loops in order to traverse arrays of objects and to post the whole set
of constraints. For instance, the last block of for loops in the
\Eclipse{} model (lines 27 to 39) has been built from the
\code{playOncePerWeek} constraint zone of the \sCOMMA{} model, but
there is two additional for loops (lines 21 and 22) since the \code{Week}
instances are contained in the \code{weeks} array. Another issue
is related to lists that cannot be accessed in the same way than
arrays in \sCOMMA{}. Thus, local variables (\code{V}$_{i}$) and the
well-known \code{nth} Prolog built-in function are introduced in the
\Eclipse{} model.


\section{Pivot metamodel and refactoring rules}\label{sec:intermediate}
The pivot model of a constraint problem is an intermediate model to be
transformed by rules. The rules may be chained to implement complex
transformations. In the following, the pivot and some structural
refactoring and optimization rules are presented.

\subsection{Pivot metamodel}
Our pivot model has been designed to support as much as possible the
constructs present in CP languages, for instance variables of
many types, data structures such as arrays, record, classes,
first-order constraints, common global constraints, and control
statements. We believe that it is better and simpler to establish a
general CP metamodel, while it is more complex to find a standard CP
concrete syntax.

\begin{figure}[tbp]
\vspace*{-4mm}
\begin{center}
  \includegraphics[width=0.90\linewidth] {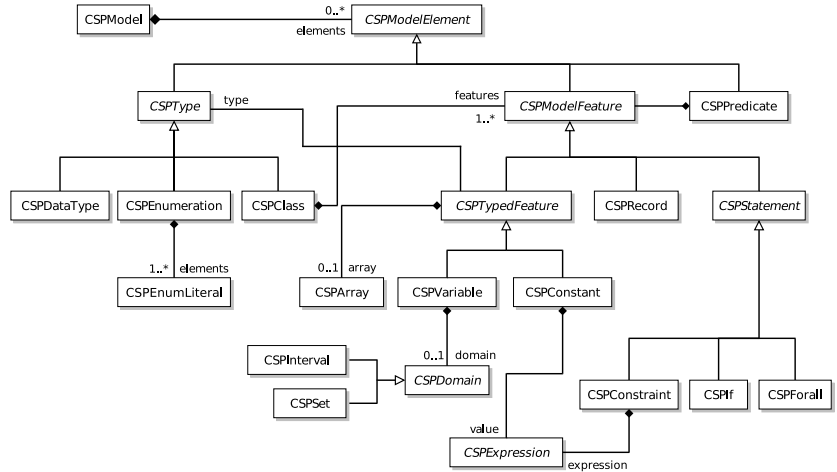}
\caption{Extract of the pivot metamodel.}\label{fig:pivot-metamodel}
\end{center}
\vspace*{-2mm}
\end{figure}

Figure~\ref{fig:pivot-metamodel} depicts the metamodel associated to
pivot models. A pivot model is composed of a collection of elements,
divided in three main concepts: types, features and the concrete
concept of predicate. The inheritance tree of types is the same as
in the \sCOMMA{} metamodel (see Figure~\ref{fig:scomma-metamodel}). The
inheritance tree for model features is also quite similar, except for
the concept of record which is an untyped collection of features.

\subsection{Pivot model refactoring}\label{sec:refactoring}
We define several refactoring steps on pivot models in order to reduce
the possible gap between source and target model. These steps are
implemented in several model transformations, most of them being
independent from the others. The idea is to refine and optimize models in order
to fit the target languages supported concepts.

Model transformations are implemented in the declarative
transformation rule language ATL~\cite{Jouault2006ATL}. This rule
language is based on a typed description of models to be processed, namely
their metamodel. In this way, rules are able to clearly state how
concepts from source metamodels are mapped to concepts from the target
ones. For the sake of simplicity, only a few of the more representative
rules of transformations are shown. ATL helpers are not
detailed, but they only consist of OCL navigation.

\subsubsection{Composition flattening}
This refactoring step replaces object variables by duplicating elements
defined in their class definition. Names of duplicated variables are
prefixed using their container name in order to avoid naming
ambiguities. This refactoring step processes object variables and their
occurrences, while other entities are copied without
modification. In fact, two ATL transformations are defined to ease each
refactoring step. The first one removes classes and object variables by
replacing them by the concept of record (see
Figure~\ref{fig:CompClass}). It can be highlighted that there is no
ATL rule where the source pattern matches elements being instances of
\code{CSPClass}. Thus, they are implicitly removed from models
(obviously no rule creates class instances). The second transformation
removes records to get flattened variables (see
Figure~\ref{fig:CompRec}).

\begin{figure}[htpb]
\begin{lstlisting}
rule Model {
  from
    s : Pivot!CSPModel
  to
    t : Pivot!CSPModel (
      name <- s.name,
      elements <- s.elements
}
rule Variable {
  from
    s : PivotCSP!CSPVariable (
      not s.mustBeDuplicated
    )
    to
      t : PivotCSP!CSPVariable (
        name <- s.name,
        type <- s.type,
        domain <- s.domain,
        isSet <- s.isSet,
        array <- s.array
      )
}
rule Variable2Record {
  from
    s : PivotCSP!CSPVariable (
      s.isObject
    )
    to
      t : PivotCSP!CSPRecord (
        name <- s.name,
        array <- s.array,
        elements <- s.type.features->collect(f|
          thisModule.duplicate(f)
        )
      )
}
\end{lstlisting}
\caption{An extract of ATL rules used to remove the concept of class in
  pivot models.\label{fig:CompClass}}
\end{figure}

In Figure~\ref{fig:CompClass}, the first rule (lines 1 to 8) is used to copy the root concept of
model. Most of other concepts are duplicated with similar rules like the
the second one (lines 9 to 22). The helper \code{mustBeDuplicated}  is
defined for each \code{CSPModelFeature}  and it returns true when: (1)
the considered element is an object variable (its type is a class) or
(2) it is a feature of a class. Using the last rule, object variables
are replaced by records. The helper \code{isObject} returns true only
if the type of variables is a class. In this rule, features of
variable classes are browsed using OCL navigation (\code{collect}
statement over \code{s.type.features}). The rule \code{duplicate} is
applied on each feature. This rule is lazy and abstract. It is
specialized for each \code{CSPModelFeature} concrete sub-concepts and
it creates as many features as it is called.

The second transformation processes records by replacing them by their
set of elements. This is easily done by collecting their elements from
their container as shown on Figure~\ref{fig:CompRec} at lines 7 to
11. The helper \code{getAllElements} returns the set of
\code{CSPModelFeature} within a record or a hierarchy of records.
\begin{figure}[htpb]
\begin{lstlisting}
rule CSPModel {
  from
    s : PivotCSP!CSPModel
  to
    t : PivotCSP!CSPModel (
      name <- s.name,
      elements <- s.elements->union(s.elements->select(r|
          r.oclIsTypeOf(PivotCSP!CSPRecord)
        )->collect(r|
          r.getAllElements
        )->flatten())
    )
}
rule RecordArray {
  from
    s : PivotCSP!CSPRecord (
      (not s.array.oclIsUndefined()) and
      s.elements->select(e|
        e.oclIsKindOf(PivotCSP!CSPStatement)
      )->size()>0
    )
    to
      t : PivotCSP!CSPForall (
        index <- i,
        constraints <- s.elements->reject(e|
            e.oclIsKindOf(PivotCSP!CSPTypedElement)
          ),
        i : PivotCSP!CSPIndexVariable (
          name <- s.name,
          domain <- d
        ),
        d : PivotCSP!CSPIntervalDomain(
          lower <- l,
          upper <- thisModule.duplicateExpr(s.array.n)
        ),
        l : PivotCSP!CSPIntVal(
          value <- 1
        )
}
\end{lstlisting}
\caption{Main ATL rules used to remove the concept of record in
  pivot models.\label{fig:CompRec}}
\end{figure}

However, some other complex rules must be defined to process
arrays of records, (formerly arrays of object variables). Indeed,
contained statements have to be encapsulated in a for loop to take
into account the constraints for all objects in the array. This task
is performed by the rule \code{RecordArray} which create a new for
loop over the record statements (lines 25 to 27). A new for loop
requires also a new index variables with its domain (lines 28 to 38).

Using the concrete syntax of \sCOMMA{}, Figure~\ref{fig:scmaComp} shows
the result of this refactoring step. The name of the variable at
line~1 corresponds to the concatenation of all object variable
names. The two for loops (lines 2 and 3) were created from the arrays
of objects using their name for index variables.
\begin{figure}[htpb]
\begin{lstlisting}[style=scma]
  int set weeks_groups_players[w*g] in[1, 9],
  forall weeks in [1,w] {
    forall groups in [1,g] {
      card(weeks_groups_players[weeks*w+groups])= g,
      ...
    }
  }
\end{lstlisting}
\caption{Extract of the social golfers pivot model after composition
  removal and enumeration removal transformations.\label{fig:scmaComp}}
\end{figure}

\subsubsection{Enumeration removal}
During this refactoring step, enumeration variables are replaced by
integer variables with a domain defined as an interval from one to the
number of elements within the enumeration. Line 1 in
Figure~\ref{fig:scmaComp} shows the result of this transformation on
the enumeration called \code{Name} in the social golfers model: 
the variable has an integer domain from 1 to 9 replacing the set of nine values \code{$\{a,b,c,d,e,f,g,h,i\}$}.
In the same way, occurences of \code{CSPEnumLiteral} are
replaced by their position in the sequence of elements of the
enumeration type.


\subsubsection{Other implemented refactoring steps}
Some other generic refactoring steps have been implemented in ATL to
handle some structural needs. They are not detailed since their
complexity is similar to the previous examples and to detail all of
them is not the scope of this paper.
\begin{itemize}
\item If statements can be replaced by one constraint based on one or
  two boolean implications. For instance, $if ~ a ~ then ~ b ~ else ~
  c$ becomes $(a\rightarrow b)\wedge(\neg a\rightarrow c)$.
\item Loop structures can be unrolled, i.e. the loop is replaced by
  the whole set of constraints it implicitly contains.  Within
  expressions, the iterator variable used by the loop structure is
  replaced by an integer corresponding to the current number of loop
  turns.
\item Expressions can be simplified if they are constants. Boolean and
  integer expressions are replaced by their evaluation. Real
  expressions are not processed, because of real number rounding
  errors. More subtle simplifactions can be performed on boolean
  expressions such as $a \vee \neg a$ that is always true. Only atomic
  boolean elements are processed by this last step.
\item Matrices are not allowed in all CP language, thus they can be
  replaced by one dimension arrays. Their occurrences in expressions
  must also be   adapted: the index of the array is computed as
  follows: $m[i,j]$ becomes $m[j+(i*ncols)]$, where $ncols$ is the
  number of columns of the matrix $m$.
\item The \Eclipse{} language does not allow some sort of
  expressions. For instance, arrays of int sets cannot be accessed
  like other arrays with \code{`[ ]'}. Thus, an \Eclipse{} specific
  transformation processes expressions and introduces local
  variables if needed, as shown on Figure~\ref{fig:eclipse-golfers}
  with \code{V}$_{i}$ variables and \code{nth} predicate calls.
\end{itemize}

\section{Handling CP languages and transformation chains}
\label{sec:bidirectional}

In this section, we describe the whole transformation chain from a
given CP language to another language.

\subsection{Parsing CP languages}
\label{sec:injection}

The front-end of our system parses a source CP language file to get a
model representation (on which transformation rules act) matching the
concepts of the CP language  (injection phase). The back-end generates
the code in the target CP language (extraction phase)  from the model
representation. Interfacing CP languages and metamodels is implemented
by means of the TCS tool~\cite{Jouault2006TCS}. This tool allows one
to smoothly associate grammars and metamodels. It is responsible for
generating parsers of CP languages and also code generators.

Figure~\ref{fig:tcs} depicts an extract of the TCS file for \sCOMMA{}.
In a TCS file every
concrete concept must have a corresponding \emph{template} to be
matched. For instance, the \code{SCMAClass} template implements the
grammar pattern for class declarations using at the same time features
of this concept defined in the metamodel of \sCOMMA{}.
At parsing time on the \sCOMMA{} social golfers example (see
Figure~\ref{fig:scomma-golfers}, the \code{"class"} token is matched
for the week class statement. Then \code{Week} is 
processed as the \code{name} attribute (a string in the metamodel) of
a new class instance. Then the \code{"\{"} token is recognized and the
class features (the array of groups and the constraint) are processed
by implicit matchings to their corresponding templates
using the \code{features} reference. Finally the \code{"\}"} token
terminates the pattern description.
In the \code{SCMAClass} template (lines 4 to 8), several TCS keywords
are used. Here is a description of the most important keywords use in
Figure~\ref{fig:tcs}:
\begin{itemize}
\item \emph{context} defines a local symbol table.
\item \emph{addToContext} adds instances to the current symbol table.
\item \emph{refersTo} accesses to the symbol tables according to
  the given parameter (here the name) to check the existence of an
  already declared element.
\end{itemize}

\begin{figure}
\begin{lstlisting}[style=tcs]
  template SCMAModel main context
    : elements;

  template SCMAClass context addToContext
    : (isMain ? "abstractmain") (isAbstract ? "abstract") "class" name
        (isDefined(superTypes) ? "extends"
          superTypes{separator=",",refersTo=name, importContext})
	"{" [ features {separator=","} ] "}" ;
	
  template SCMAVariable addToContext
    : (isSet ? "set" : "") type {refersTo=name}
        name (isDefined(array) ? array)
          (isDefined(domain) ? "in" domain);
\end{lstlisting}
\caption{Linking the grammar and the metamodel of \sCOMMA{} with
  TCS.}\label{fig:tcs}
\end{figure}

\subsection{Model checking rules}
The presented metamodels (see section 2) and the previous subsection
show how to get CP language models. However, many irrelevant or
erroneous models can be obtained without any additional checking~\cite{Bezivin2005}. For
instance, variables may be defined with empty domains or expressions
may be ill made (e.g. several equalities in an equality constraint).

Several ATL transformations are used to check source models. We
transform a source CP model to a model conform to the metamodel
Problem defined in the ATL
zoo\footnote{\url{http://www.eclipse.org/m2m/atl/atlTransformations/\#KM32Problem}}.
A \code{Problem} model corresponds to a set of \code{Problem}
elements. This concept is only composed of three features:
\begin{itemize}
\item \code{severity} is an attribute with an enumerated type which
  possible values are: error, warning and critic.
\item \code{location} is a string used to store le location of the
  problem in the source file.
\item \code{description} is a string used to defined a relevant
  message to descibe the problem.
\end{itemize}

Multiple ATL rules have been implemented to check models. Here is an
extract of the list of properties to check:
\begin{itemize}
\item Some type checking on expressions. Operands must have a
  consistent type with the operator. For instance, an equality
  operator may operate on arithmetic expressions.
\item The consistency of variable domains : they must be based on constant
  expressions and interval domains must have a lower bound smaller
  than the upper bound.
\item No composition or inheritance loops in \sCOMMA{}.
\end{itemize}

\subsection{Chaining model transformations}\label{subsec:chain}
After the injection step or before the extraction step, models have to
be transformed with respect to our pivot metamodel.
All the refactoring steps presented in Section~\ref{sec:refactoring}
are clearly not necessary in a transformation chain.
Indeed, it clearly depends on the modeling structures of the source
and target CP languages. The idea is to use most of constructs
supported by the target language to have a target model close, in
terms of constructs, to our source model. For instance, when
translating a \sCOMMA{} model to \Eclipse{}, we should transform the
objects. So, we choose the composition flattening step.  We also need
the enumeration removal and other refactoring steps such as the use of
local variables and \code{nth} predicates. Optionally, we may select
the expression simplification steps.

The whole transformation chain is based on three kind of tasks: (1)
injection/extraction steps, (2) transformation steps from/to the pivot
model, (3) relevant refactoring steps. Transformation chains are currently
performed using
Ant scripts\footnote{http://wiki.eclipse.org/index.php/AM3\_Ant\_Tasks}.
These
scripts are hand-written, but they can be automatically generated
using the am3 tool~\cite{Barbero2008} and the concept of
megamodel~\cite{Fritzsche2009} to get a graphical interfaces to
manage terminal models, metamodels and complex transformation chains.
However, Automating the building of transformation chains is not possible with
current tools. It would require to deeply analyze models and transformations
to build relevant transformation chains.

\section{Experiments}\label{sec:experiments}
The benchmarking study was performed on a 2.66Ghz computer with 2GB
RAM running Ubuntu. The ATL regular VM is used for all model-to-model
transformations, whereas TCS achieve the text-to-model and
model-to-text tasks. Five CP problems were used to validate our
approach as shown in Table~\ref{tab:bench1}. The second column
represents the number of lines of the \sCOMMA{} source files. The next
columns correspond to the time of atomic steps (in seconds): model
injection (Inject), transformations from \sCOMMA{} to Pivot (s-to-P),
refactoring composition structures (Comp), refactoring enumeration
structures (Enum), transformations from Pivot to \Eclipse{} (P-to-E),
and target file extraction (Extract). The next column details the total time
of complete transformation chains, and the last column corresponds to
the number of lines of the generated \Eclipse{} files.

\begin{table}[htpb]
\begin{center}
\begin{scriptsize}
\begin{tabular}{|l||c|c|c|c|c|c|c|c|c|}
\hline
Problems & Lines & Inject & s-to-P & Comp & Enum &
P-to-E & Extract & Total & Lines\\
 & (-) & (s) & (s) & (s) & (s) &
 (s) & (s) & (s) & (-)\\
\hline
SocialGolfers     & 42 & 0.107 & 0.169 & 0.340 & 0.080 &
 0.025 &
    0.050 & 0.771 & 38\\
Engine            & 112 & 0.106 & 0.186 & 0.641 & 0.146 &
 0.031 &
    0.056 & 1.166 & 78\\
Send              & 16 & 0.129 & 0.160 & 0.273 & - &
 0.021 &
    0.068 & 0.651 & 21\\
StableMarriage    & 46 & 0.128 & 0.202 & 0.469 & 0.085 &
 0.027 &
    0.040 & 0.951 & 26\\
10-Queens         & 14 & 0.132 & 0.147 & 0.252 & - &
 0.017 &
    0.016 & 0.564 & 12\\
\hline
\end{tabular}
\end{scriptsize}
\end{center}
\caption{Times for complete transformation chains of several classical problems.}\label{tab:bench1}
\end{table}

The transformation chain is efficient for these small problems. The text
file injection and extraction are fast. The parsing phase is more
expensive than the extraction, since it requires the management of
symbol tables. The extraction phase settle for reading the
\Eclipse{} model. It can also be noticed that model transformations to
and from the pivot are quite efficient, more especially the
transformation to \Eclipse{} model. It can be explained by the
refactoring phases on the pivot model which simplify and reduce the
data to process.
We see that the composition flattening step is the more expensive.
In particular, the Engine problem exhibits the slowest running time,
since it corresponds to the design of an engine with more object 
compositions.

\begin{table}[htpb]
\begin{center}
\begin{scriptsize}
\begin{tabular}{|l||c|c|c|c|c|c|c|c|c|c|}
\hline
Problems & Inject & s-to-P & Comp &
 Forall & P-to-E & Extract & Total & Lines & Total/Lines\\
 & (s) & (s) & (s) &
 (s) & (s) & (s) & (s) & (-) & (-)\\
\hline
5-Queens      & 0.132 & 0.147 & 0.252 &
 0.503 & 0.071 & 0.019 &
                1.124 & 80 & $\approx$0.014\\
10-Queens     & 0.132 & 0.147 & 0.252 &
 1.576 & 0.280 & 0.060 &
                2.447 & 305 & $\approx$0.008\\
15-Queens     & 0.132 & 0.147 & 0.252 &
 3.404 & 0.659 & 0.110 &
                4.704 & 680 & $\approx$0.007\\
20-Queens     & 0.132 & 0.147 & 0.252 &
 6.274 & 1.224 & 0.178 &
                8.207 & 1205 & $\approx$0.006\\
50-Queens     & 0.132 & 0.147 & 0.252 &
 32.815 & 13.712 & 1.108 &
                48.166 & 7505 & $\approx$0.006\\
75-Queens     & 0.132 & 0.147 & 0.252 &
 80.504 & 54.286 & 2.456 &
                137.777 & 16880 & $\approx$0.008\\
100-Queens    & 0.132 & 0.147 & 0.252 &
 175.487 & 126.607 & 4.404 &
                307.029 & 30005 & $\approx$0.010\\
\hline
\end{tabular}
\end{scriptsize}
\end{center}
\caption{Time of complete transformation chains of the N-Queens problem.}\label{tab:bench2}
\vspace*{0mm}
\end{table}

Table~\ref{tab:bench2} presents seven different sizes of
the N-Queens problem where the loop unrolling step has been
applied. This experiment allows us to check the scalability of our
approach according to model sizes. It can be analyzed 
through the ratio given in the last column which aims at quantifying 
the efficiency of a transformation chain considering the 
execution time per generated lines. 

As shown on this table, the ratio first decreases, but after 50-Queens it
slowly grows up. In fact, the first four row ratios are impacted by the
steps before the loop unrolling process, but for the last three rows they
become neglectible comparing to the whole execution time.
It may be noticed that for big problems (after 50-Queens) the ratio
smoothly increases. We can thus conclude that our approach is applicable 
even for huge models, although translations times are not the major 
concerns in CP.

\section{Conclusion and Future Work}

In this paper, we propose a new framework for constraint model transformations. 
This framework is supported by a set of MDE tools that allow an easy 
design of translators to be used in the whole transformation chain. 
This chain is composed by three main steps: from the source to the pivot 
model, refining of the pivot model and from the pivot model to the 
target. The hard transformation work (refactoring/optimization) is 
always performed by the pivot which provide reusable and flexible
transformations. The transformations from/to pivot become 
simple, thus facilitating the integration of new language
transformations. In this paper, only two languages are presented, but
translation processes with Gecode and Realpaver~\cite{GranvilliersACM2006} are already implemented.

In a near future, we intend to increase the number of CP languages our
approach supports. We also want to define more pivot refactoring
transformations to optimize and restructure models. 
Another major outline for future work is to improve the management of
complex CP models transformation chains. Models can be qualified to
determine their level of structure and to automatically choose the
required refactoring steps according to the target language.

\bibliographystyle{plain}
\bibliography{atl_2009}

\begin{thebibliography}{10}

\bibitem{Barbero2008}
M.~Barbero, F.~Jouault, and L.~B\'ezivin.
\newblock Model driven management of complex systems: Impementing the
  macroscope's vision.
\newblock In {\em 15th International Conference on Engineering of
  Computer-Based Systems}, 2008.

\bibitem{Bezivin2005}
J.~B\'ezivin and F.~Jouault.
\newblock Using atl for checking models.
\newblock In {\em Proceedings of the International Workshop on Graph and Model
  Transformation (GraMoT)}, Tallinn, Estonia, 2005.

\bibitem{Brand2008}
S.~Brand, G.~J. Duck, J.~Puchinger, and P.~Stuckey.
\newblock Flexible, rule-based constraint model linearisation.
\newblock In P.~Hudak and D.~Warren, editors, {\em Practical Aspects of
  Declarative Languages}, volume 4902 of {\em LNCS}, pages 68--83. Springer,
  2008.

\bibitem{Chenouard2008}
R.~Chenouard, L.~Granvilliers, and R.~Soto.
\newblock Model-driven constraint programming.
\newblock In {\em ACM SIGPLAN PPDP}, pages 236--246, Valencia, Spain, 2008.

\bibitem{FrischIJCAI2007}
A.~M. Frisch, M.~Grum, C.~Jefferson, B.~Mart\'inez Hern\'andez, and I.~Miguel.
\newblock The design of essence: A constraint language for specifying
  combinatorial problems.
\newblock In {\em IJCAI}, pages 80--87, 2007.

\bibitem{FrischIJCAI2005}
A.M. Frisch, C.~Jefferson, B.~Martinez-Hernandez, and I.~Miguel.
\newblock {The Rules of Constraint Modelling}.
\newblock In {\em IJCAI 2005}, pages 109--116, Edinburgh, Scotland, 2005.

\bibitem{Fritzsche2009}
M.~Fritzsche, H.~Bruneliere, B.~Vanhooff, Y.~Berbers, F.~Jouault, and
  W.~Gilani.
\newblock Applying megamodelling to model-driven performance engineering.
\newblock In {\em 16th Annual IEEE ECBS}, San Fransisco, USA. April 13-16,
  2009.

\bibitem{CHRBook2009}
T.~Fr{\"u}hwirth.
\newblock {\em {Constraint Handling Rules}}.
\newblock Cambridge University Press, June 2009.
\newblock to appear.

\bibitem{GranvilliersACM2006}
L.~Granvilliers and F.~Benhamou.
\newblock Algorithm 852: Realpaver: an interval solver using constraint
  satisfaction techniques.
\newblock {\em ACM Trans. Math. Softw.}, 32(1):138--156, 2006.

\bibitem{JouaultATL2008}
F.~Jouault, F.~Allilaire, J.~B\'ezivin, and I.~Kurtev.
\newblock Atl: A model transformation tool.
\newblock {\em Science of Computer Programming}, 72(1-2):31 -- 39, 2008.
\newblock Special Issue on Second issue of experimental software and toolkits
  (EST).

\bibitem{Jouault2006TCS}
F.~Jouault, J.~B{\'e}zivin, and I.~Kurtev.
\newblock {TCS: a DSL for the specification of textual concrete syntaxes in
  model engineering.}
\newblock In {\em Conference on Generative Programming and Component
  Engineering (GPCE 2006)}, pages 249--254, 2006.

\bibitem{Jouault2006ATL}
F.~Jouault and I.~Kurtev.
\newblock {Transforming Models with ATL.}
\newblock In {\em MoDELS Satellite Events 2005}, volume 3844 of {\em LNCS},
  pages 128--138. Springer, 2005.

\bibitem{Nethercote2007}
N.~Nethercote, P.~J. Stuckey, R.~Becket, S.~Brand, G.~J. Duck, and G.~Tack.
\newblock Minizinc: Towards a standard cp modelling language.
\newblock In C.~Bessi{\`e}re, editor, {\em 13th International CP Conference},
  volume 4741 of {\em LNCS}, pages 529--543. Springer, 2007.

\bibitem{OMG_MDA}
OMG.
\newblock {\em Model Driven Architecture (MDA) Guide V1.0.1, 2003}.
\newblock http://www.omg.org/cgi-bin/doc?omg/03-06-01.

\bibitem{PugetCP04}
J.F. Puget.
\newblock Constraint programming next challenge: Simplicity of use.
\newblock In {\em CP 2004}, LNCS 3258, pages 5--8, 2004.

\bibitem{SotoICTAI2007}
R.~Soto and L.~Granvilliers.
\newblock The design of comma: An extensible framework for mapping constrained
  objects to native solver models.
\newblock In {\em IEEE ICTAI 2007}, pages 243--250, 2007.

\bibitem{Wallace97eclipse}
M.~Wallace, S.~Novello, and J.~Schimpf.
\newblock Eclipse: A platform for constraint logic programming, 1997.

\end{thebibliography}
\end{document}